%% file: main.tex
\documentclass{article}
\usepackage[preprint]{neurips_2025_custom}
\usepackage{fix-cm}

\usepackage{xcolor}         
\definecolor{linkColor}{rgb}{0.2,0.4,0.6}
\definecolor{myblue}{HTML}{0379AC}
\definecolor{myred}{HTML}{A50E50}
\usepackage[utf8]{inputenc} 
\usepackage[T1]{fontenc}    
\usepackage[colorlinks=true,linkcolor=linkColor,citecolor=linkColor,filecolor=linkColor,urlcolor=linkColor]{hyperref}       
\usepackage{url}            
\usepackage{booktabs}       
\usepackage{amsfonts}       
\usepackage{nicefrac}       
\usepackage{microtype}      
\usepackage{graphicx}
\usepackage{arydshln}
\usepackage{booktabs}
\usepackage{multirow}
\usepackage{caption}
\usepackage{subcaption}
\usepackage{makecell}
\usepackage{csquotes}
\usepackage{epigraph}
\RequirePackage{algorithm}
\RequirePackage{algorithmic}

\input{settings.tex}
\input{math_commands.tex}

\newcommand\ours{\textsc{RPT}}
\newcommand\rpt{reinforcement pre-training}
\newcommand\RPT{Reinforcement Pre-Training}

\title{Reinforcement Pre-Training}

\author{
Qingxiu Dong\thanks{~Equal contribution. $\diamond$ Contact person: \href{mailto:fuwei@microsoft.com}{fuwei@microsoft.com}.}$~~^{\dag\ddag}$~~~~~~Li Dong\footnotemark[1]$~~^{\dag}$ \\
~\bf Yao Tang$^{\dag}$~~~~~~Tianzhu Ye$^{\dag\S}$~~~~~~Yutao Sun$^{\dag\S}$~~~~~~Zhifang Sui$^{\ddag}$~~~~~~Furu Wei$^{\dag}$$^{\diamond}$ \\
~$^\dag$ Microsoft Research \\
~$^\ddag$ Peking University \\
~$^\S$ Tsinghua University \\
~{\href{https://aka.ms/GeneralAI}{https://aka.ms/GeneralAI}}
}

\begin{document}

\maketitle

\begin{abstract}
In this work, we introduce \textbf{Reinforcement Pre-Training} (RPT) as a new scaling paradigm for large language models and reinforcement learning (RL). Specifically, we reframe next-token prediction as a reasoning task trained using RL, where it receives verifiable rewards for correctly predicting the next token for a given context. RPT offers a scalable method to leverage vast amounts of text data for \textbf{general-purpose RL}, rather than relying on domain-specific annotated answers. By incentivizing the capability of next-token reasoning, RPT significantly improves the language modeling accuracy of predicting the next tokens. Moreover, RPT provides a strong pre-trained foundation for further reinforcement fine-tuning. The scaling curves show that increased training compute consistently improves the next-token prediction accuracy. The results position RPT as an effective and promising scaling paradigm to advance language model pre-training.
\end{abstract}

\vfill{}

\begin{figure}[h]
\centering
\includegraphics[width=\linewidth]{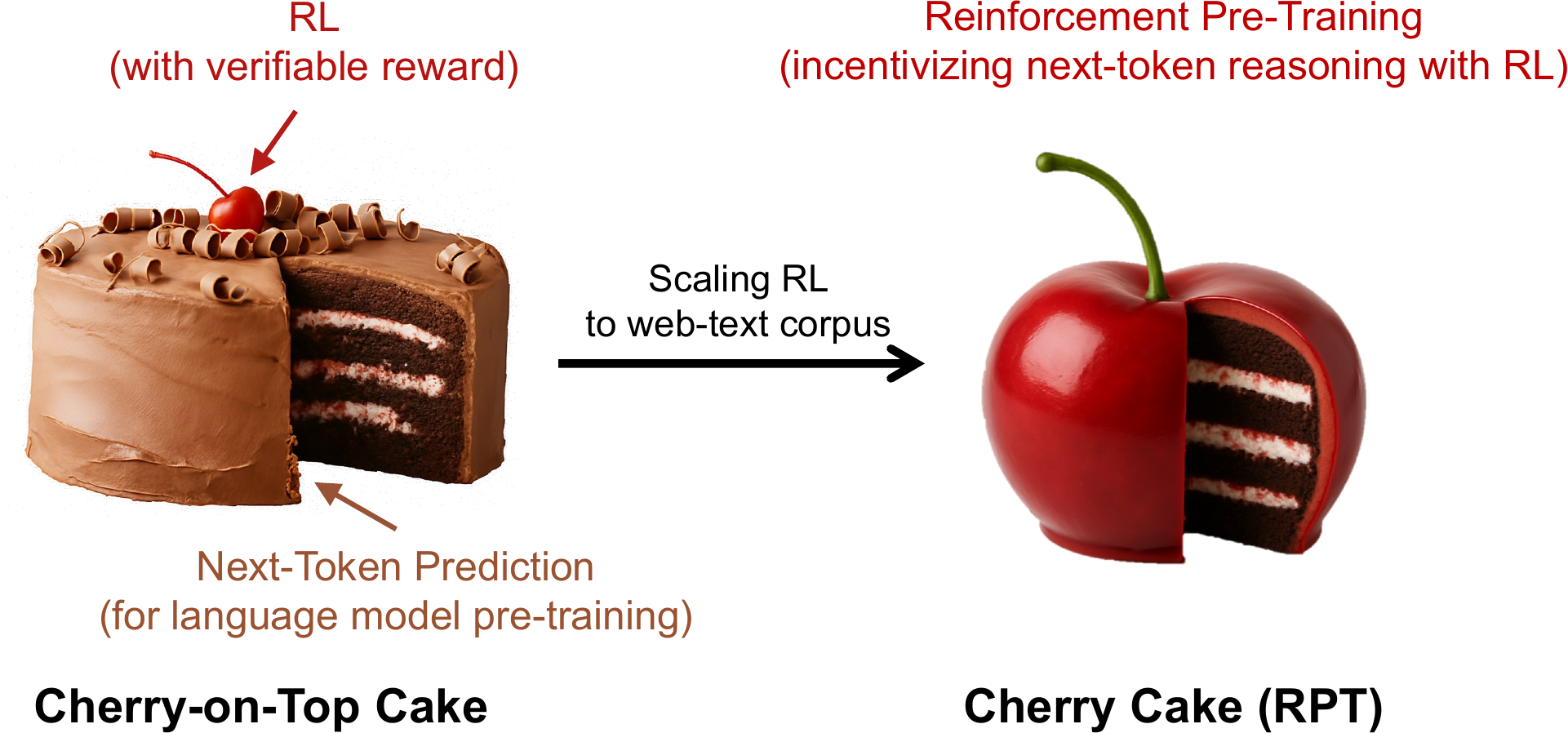}
\caption{Reinforcement pre-training (RPT) reframes next-token prediction as a reasoning task, where the language model is incentivized via reinforcement learning (RL) to reason about and correctly predict the next token. The proposed approach allows RL to be scaled to the web-text corpus.
The image of the cherry-on-top cake is taken from LeCun's slides \cite{lecun:cake}.
}
\label{fig:cherry}
\end{figure}

\vfill{}

\newpage

\section{Introduction}
\label{sec:intro}

Large language models (LLMs) have demonstrated remarkable capabilities across a wide range of tasks, largely driven by the scalability of the next-token prediction objective on vast text corpora. This self-supervised paradigm has proven to be an effective general-purpose pre-training approach. Concurrently, reinforcement learning (RL) has emerged as a powerful technique for fine-tuning LLMs, aligning them with human preferences or enhancing specific skills such as complex reasoning \cite{instructgpt, o1, deepseekr1}.

However, current applications of RL in LLM training face scalability and generality challenges. Reinforcement learning from human feedback \cite{instructgpt}, while effective for alignment, relies on costly human preference data, and its learned reward models can be susceptible to reward hacking, limiting scalability. Alternatively, reinforcement learning with verifiable rewards (RLVR) \cite{tulu3} utilizes objective, rule-based rewards, often from question-answer pairs. While this mitigates reward hacking, RLVR is typically constrained by the scarcity of annotated data with verifiable answers, restricting its application to domain-specific fine-tuning rather than general-purpose pre-training.

In this work, we introduce \rpt{} (\ours{}), a novel paradigm that bridges the gap between scalable self-supervised pre-training and the power of reinforcement learning. \ours{} reframes the fundamental next-token prediction task as a next-token reasoning process. For any given context in a pre-training corpus, the model is incentivized to reason about the subsequent token before predicting it. It receives a verifiable, intrinsic reward based on the correctness of its prediction against the ground-truth next token from the corpus itself. This approach transforms the vast, unannotated text data typically used for next-token prediction into a massive dataset for general-purpose RL, without requiring external annotations or domain-specific reward functions.

This approach offers several crucial advantages.
First, \ours{} is inherently scalable and general-purpose: it leverages the same vast, unannotated text data used for standard next-token prediction, transforming it into a massive dataset for general-purpose RL without requiring external annotations.
Second, the use of direct, rule-based reward signals (i.e., the correctness of the predicted next token) inherently minimizes the risk of reward hacking often associated with complex, learned reward models.
Third, by explicitly encouraging next-token reasoning patterns, \ours{} promotes deeper understanding and generalization instead of merely memorizing next tokens. The model learns to explore and validate hypotheses about why a certain token should follow, fostering more robust representations.
Finally, the internal reasoning process during pre-training effectively allows the model to allocate more ``thought'' or computational effort to each prediction step, akin to a form of inference-time scaling applied at training time for each token, which directly contributes to improved next-token prediction accuracy.

Our experiments demonstrate that \ours{} significantly improves the accuracy of predicting next tokens.
\ours{} also provides a more robust pre-trained foundation for subsequent reinforcement fine-tuning, leading to better final task performance. The scaling curves reveal that increased training compute under the \ours{} framework consistently improves next-token prediction accuracy, indicating its potential as a sustainable scaling strategy. These results position \rpt{} as an effective and promising new paradigm to advance the pre-training of large language models.

Our contributions are summarized as follows:
\begin{itemize}[leftmargin=*]
\setlength\itemsep{0.01em}
\item We introduce \rpt{} (\ours{}), a new scaling paradigm that reframes next-token prediction as a reasoning task trained with reinforcement learning, utilizing intrinsic verifiable rewards derived directly from the pre-training corpus.
\item \ours{} offers a scalable and general-purpose approach to RL pre-training, minimizing reward hacking through rule-based rewards and promoting generalization by encouraging next-token reasoning patterns over rote memorization.
\item \ours{} significantly improves next-token prediction accuracy and exhibits favorable scaling properties, where performance consistently improves with increased training compute.
\item \ours{} yields a stronger pre-trained foundation for subsequent reinforcement fine-tuning and enhances zero-shot performance on various downstream tasks.
\end{itemize}

\section{Preliminary}
\paragraph{Next-Token Prediction (NTP)}
Next-token prediction is the fundamental training objective for modern large language models~\cite{achiam2023gpt-4}. Given an input sequence $x_0 \cdots x_{T}$ from the training corpus, the model is trained to maximize the following objective:
\begin{equation}
\label{eq:ntp}
\mathcal{J}_\text{NTP}(\theta) = \sum_{t=1}^{T} \log P(x_{t} \mid x_0, x_1, \ldots, x_{t-1}; \theta),
\end{equation}
where $\theta$ represents the parameters of the language model.

\paragraph{Reinforcement Learning with Verifiable Rewards (RLVR)}
RLVR employs a reinforcement learning objective to enhance specific skills with verifiable answers~\cite{tulu3}.
RLVR requires a labeled dataset of question-answer pairs $\mathcal{D} = \{(q, a)\}$. For a specific pair $(q,a) \in \mathcal{D}$, the LLM $\pi_\theta$ generates a response $o\sim\pi_\theta(\cdot\mid q)$. A deterministic verifier $\mathcal{V}$ calculates a verifiable reward $r = \mathcal{V}(o, a)$, and the model is trained to maximize the expected reward:

\begin{equation}
\label{eq:rlvr}
\mathcal{J}_\text{RLVR}(\theta) = \mathbb{E}_{(q,a)\sim\mathcal{D},\ o\sim\pi_\theta(\cdot\mid q)}\left[r(o, a)\right].
\end{equation}

\section{Reinforcement Pre-Training}
\label{sec:rpt}

\subsection{Pre-Training Task: Next-Token Reasoning}
\label{sec:ntr}

\begin{figure}
\centering
\includegraphics[width=0.96\linewidth]{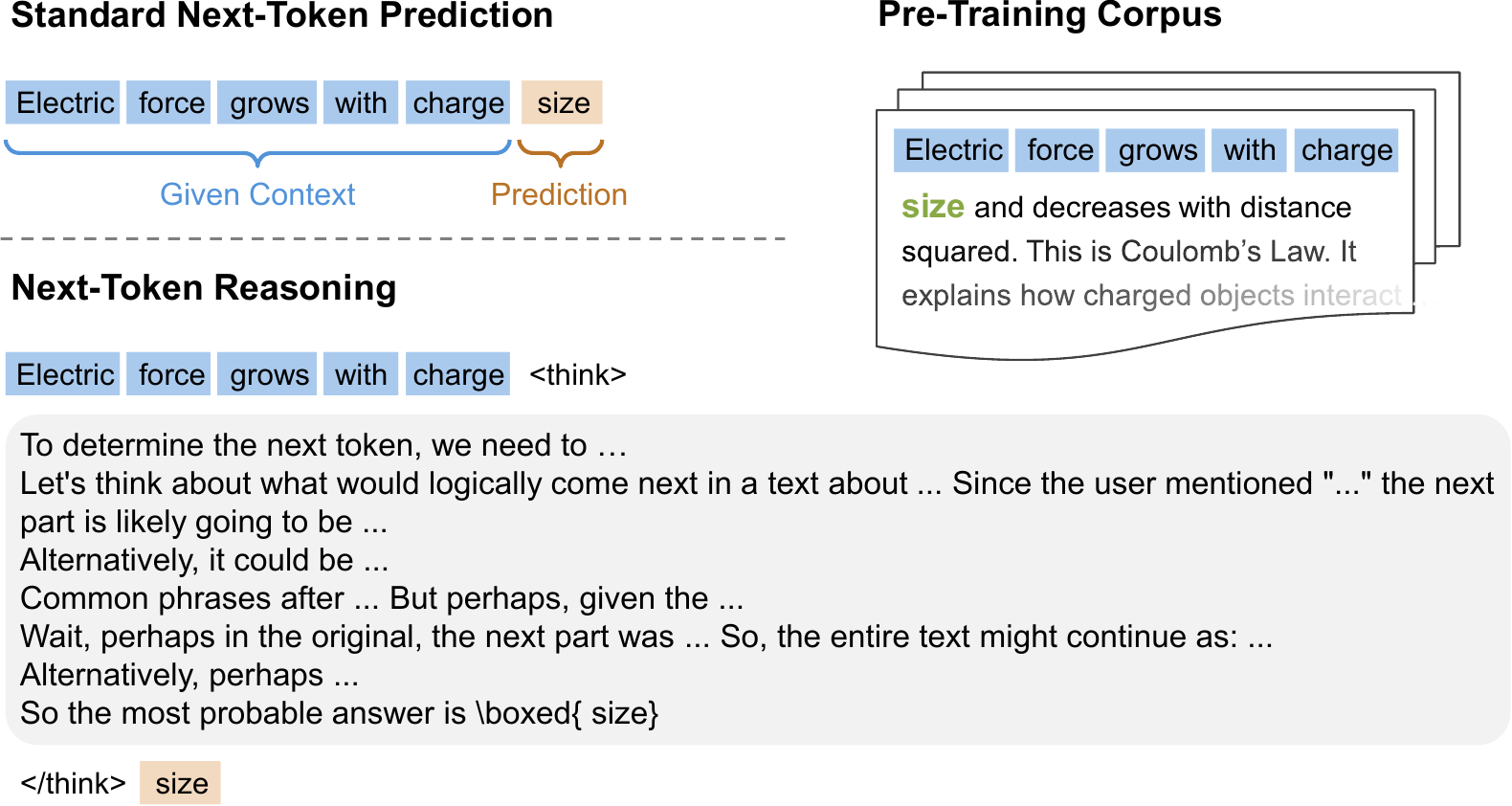}
\caption{Comparison of standard next‑token prediction and next‑token reasoning. Standard next‑token prediction estimates the next token in the pre-training corpus directly, while next‑token reasoning performs reasoning over multiple tokens before making the prediction.}
\label{fig:next-token-reason}
\end{figure}

We propose the next-token reasoning task for language modeling.
Given an input sequence $x_0 \cdots x_{T}$ from the training corpus, for each position $t \in \{1, \ldots, T \}$, the prefix  $x_{<t}$ is treated as the context, and ground-truth next token is $x_t$.
In the next-token reasoning task, the model $\pi_{\theta}$ is required to generate a chain-of-thought reasoning sequence, denoted by $c_t$,  
before generating a prediction $y_t$ for the next token. 
The overall model response is $o_t = (c_t, y_t)$,  
$o_t \sim \pi_{\theta}(\cdot \mid  x_{<t})$.

As illustrated in \Cref{fig:next-token-reason},
the long chain-of-thought process for next-token reasoning can involve various reasoning patterns such as brainstorming, self-critique and self-correction. 
The next-token reasoning task reconstructs the pre-training corpus into a vast set of reasoning problems, shifting pre-training beyond learning superficial token-level correlations to understanding the hidden knowledge behind them and making RL scaling possible.

\subsection{Pre-Training with Reinforcement Learning}

\begin{figure}
\centering
\includegraphics[width=\linewidth]{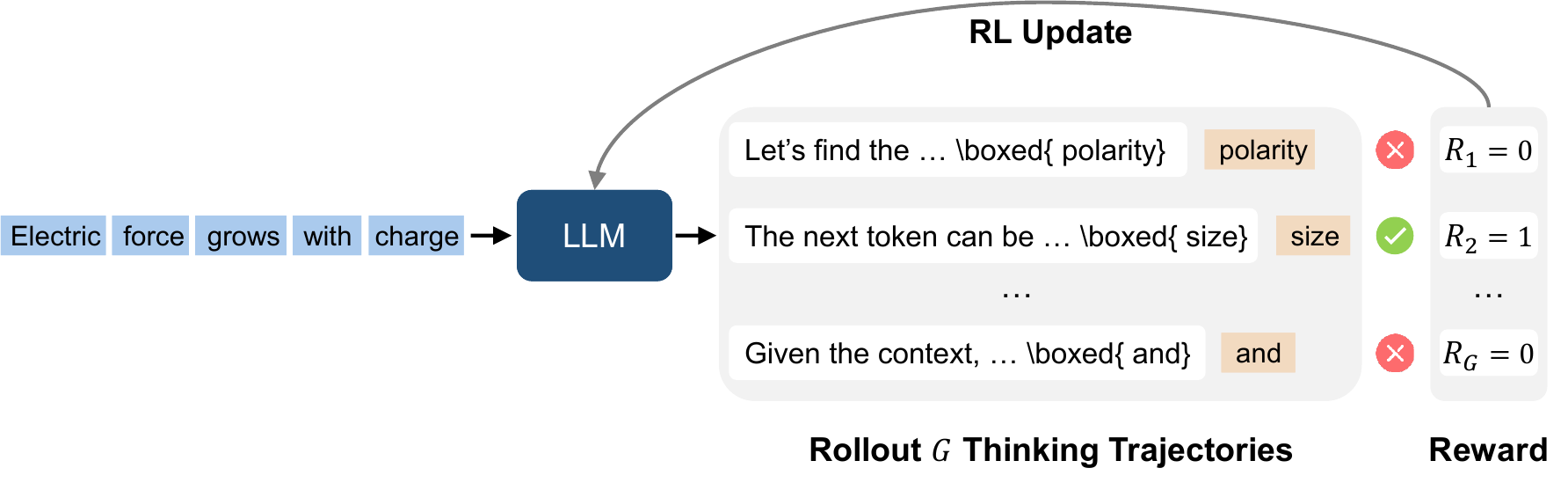}
\caption{An illustration of \rpt{}. Given a context with a missing continuation, the LLM performs on-policy rollouts to generate $G$ different thinking trajectories. Each includes an intermediate reasoning step and a final prediction for the next token. A positive reward is assigned if the prediction matches the ground-truth token; otherwise, the reward is zero. This reward signal is used to update the LLM, encouraging trajectories that lead to accurate continuations.}
\label{fig:rpt}
\end{figure}

Reinforcement pre-training (\ours{}) trains LLMs to perform next-token reasoning via on-policy reinforcement learning, as illustrated in \Cref{fig:rpt}.
For the context $x_{<t}$, we prompt the language model $\pi_{\theta}$ to generate $G$ responses (thinking trajectories), $\{o^{i}_t\}_{i=1}^{G}$. Each response $o^i_t = (c^i_t, y^i_t)$ consists of a chain-of-thought reasoning sequence $c^i_t$ and a final prediction sequence $y^i_t$.

To verify the correctness of $y^i_t$, we introduce a \textbf{prefix matching reward}, which supports verifying predictions that span multiple tokens or involve out-of-vocabulary tokens.\footnote{Additional reward design choices for next-token reasoning are discussed in \Cref{app:reward}.}
Let $\overline{x}_{\geq t}$ and $\overline{y}^i_t$ denote the byte sequences of the ground-truth completion sequence $x_{\geq t}$ and the prediction $y^i_t$, respectively. Denote the byte length of $\overline{y}^i_t$ by $l$. We define the cumulative byte lengths of the tokens in the ground-truth completion sequence as valid boundaries, and denote this set by $\mathcal{L}_{gt}$.
Formally, the reward $r^{i}_t$ for the $i$-th output for $x_{<t}$ is defined as:
\begin{equation}
r^{i}_t = \begin{cases} 1 & \text{if } \overline{y}^i_t = \overline{x}_{\geq t}[1:l] \text{ and } l \in \mathcal{L}_{gt} \\ 0 & \text{otherwise} \end{cases},
\end{equation}
where the reward is 1 if the byte sequence of the prediction is an exact prefix of the ground-truth completion sequence and its length $l$ matches any valid token boundary.

Let $\mathcal{D}$ be the set of all $\{x_{<t}\}_{t=1}^{T}$, the model is trained to maximize the expected reward:

\begin{equation}
\label{eq:rpt}
\mathcal{J}_\text{\ours{}}(\theta) = \mathbb{E}_{(x_{<t},x_{\geq t})\sim\mathcal{D},\ \{o^{i}_t\}_{i=1}^{G} \sim\pi_\theta(\cdot\mid x_{<t})}\left[r^{i}_t\right].
\end{equation}

\subsection{Pre-Training Setup}
\label{sec:method:setup}

We use the OmniMATH dataset~\cite{omnimath} for \rpt{}. OmniMATH contains 4,428 competition-level mathematical problems and solutions from official websites such as AoPS Wiki\footnote{\url{https://artofproblemsolving.com/wiki/index.php}} and AoPS forum\footnote{\url{https://artofproblemsolving.com/community/c13_contests}}.
Since many tokens are easily predictable even without reasoning, we perform token-level data filtering before \rpt{}. 
Particularly, we use Deepseek-R1-Distill-Qwen-1.5B as a small proxy model.
For each token, we calculate the proxy model entropy on the top-16 next tokens. 
By applying an entropy threshold, we filter out low-entropy positions, prioritizing training on challenging tokens that require greater computational effort to predict.

In all experiments, we use Deepseek-R1-Distill-Qwen-14B~\cite{deepseekr1} as the base model.
R1-Distill-Qwen-14B serves as a good starting point for reinforcement learning due to its basic reasoning capabilities. 
We implement our training framework with the verl library~\cite{verl} and use vllm for inference. 
We employ the GRPO algorithm~\cite{deepseekr1}, with specific hyperparameters detailed in \Cref{app:hp}. 
During training, we adopt an 8k training length, a learning rate of $1 \times 10^{-6}$, zero KL penalty, and a batch size of 256 questions. For each question, G=8 responses are sampled, and for the rollout process, we use a temperature of 0.8.
From each response, we directly extract the full sequence inside the last \textbackslash boxed\{\} following the special token `</think>' as the model prediction for the next token.
Starting from 500 steps, we utilize dynamic sampling to boost training efficiency~\cite{dapo}. The total training steps for our main experiment is 1,000.
The prompt template and its variants are discussed in \Cref{app:templates}.

\subsection{Evaluation of Pretrained Models}
\label{sec:method:eval}

Once the model is pretrained, we can directly conduct next-token prediction and reinforcement fine-tuning on downstream tasks.
We use the settings to show that \rpt{} improves the language modeling capabilities and reasoning abilities of large language models.

\paragraph{Language Modeling}
Given the next-token reasoning objective, our models can be naturally used for language modeling.
We report the next-token prediction accuracy to evaluate the language modeling performance and scaling properties of \ours{}.

\paragraph{Reinforcement Fine-Tuning on Downstream Tasks}
We conduct continual RL fine-tuning with \ours{} models in a pretrain-then-finetune manner.
Since \ours{} aligns the pre-training process with RL, the objective gap between pre-training and RL during post-training is minimized. We evaluate whether the \rpt{} process further enhances post-training on end tasks.

\section{Experiments}
\label{sec:exp}

\subsection{Language Modeling}
\label{sec:explm}
We evaluate the language modeling performance on a held-out validation set of 200 samples from OmniMATH. Following the entropy-based data filtering strategy described in our setup (\Cref{sec:method:setup}), we categorize token positions in the validation set according to their difficulty. Specifically, we calculate the entropy at each token position using R1-Distill-Qwen-14B. We then designate positions as belonging to easy, medium, or hard splits if their entropy exceeds thresholds of 0.5, 1.0, and 1.5, respectively.
For comparison, we report the performance of R1-Distill-Qwen-14B evaluated in two different ways: (1) Standard next-token prediction, selecting the token with the highest probability; and (2) Next-token reasoning, generating a chain-of-thought before the final prediction. We also include the results of Qwen2.5-14B, as it is the base model for R1-Distill-Qwen-14B.

\input{figure/lm_ntp.tex}

As shown in \Cref{tab:ntp:acc}, \ours{}-14B achieves consistently higher next-token prediction accuracy across all difficulty levels compared to R1-Distill-Qwen-14B. 
Remarkably, it matches the performance of a significantly larger model, i.e., R1-Distill-Qwen-32B (\Cref{fig:lm-avg}). These results suggest that \rpt{} is effective at capturing the complex reasoning signals underlying token generation, and holds strong potential for improving the language modeling capability of LLMs.

\subsection{Scaling Properties of \RPT{}}
\label{sec:scaling}
In this section, we investigate the scaling properties of \rpt{}.
The loss achieved by next-token pre-training on natural language corpus empirically follows a power-law decay with respect to model size, number of training tokens, and training compute~\cite{scalinglaw,scaling:law}. Below, we analyze the scaling behavior of \ours{} specifically with respect to training compute \(C\). We model this relationship using the following power-law form: 
\begin{equation}
P(C) = \frac{A}{C^{\alpha}} + P^{*}
\label{eq:scaling_rpt}
\end{equation}
where \(P(C)\) denotes the next-token prediction accuracy on the validation set. $P^{*}$, $\alpha$, and $A$ are parameters to be estimated.

We evaluate the next-token prediction accuracy of \ours{} at various training steps (100, 200, 400, 800, 1000, and 1200) and convert them into the corresponding training compute. 
To assess the impact of data difficulty, we consider validation splits filtered by entropy thresholds 0.5 (easy), 1.0 (medium), and 1.5 (hard). A higher threshold corresponds to more challenging inputs for the LLM. For each difficulty level, we fit the results according to \Cref{eq:scaling_rpt}. We measure the goodness of fit using the coefficient of determination $R^2$, which quantifies how well the scaling curve fits the observed data. 

\begin{figure}
\centering
\includegraphics[width=0.5\linewidth]{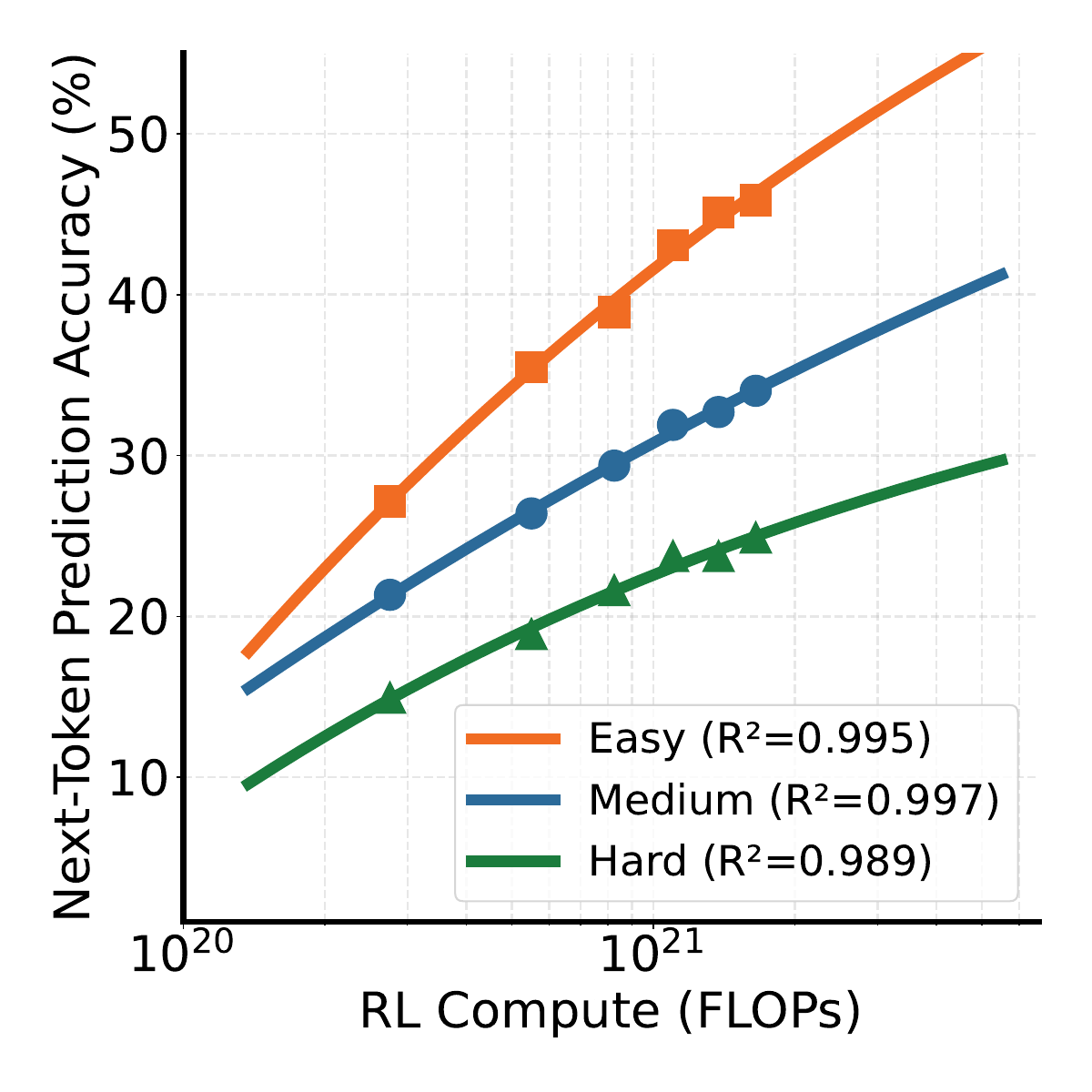}
\caption{Next-token prediction accuracy of \rpt{} improves consistently with increased training compute under all data difficulties. The fitted curves exhibit high coefficients of determination, indicating alignment between the predicted and observed values.}
\label{fig:scaling}
\end{figure}

As shown in \Cref{fig:scaling}, the next-token prediction accuracy of \ours{} improves reliably as the training compute is scaled up.
High $R^2$ values across all difficulty levels demonstrate that the fitted curves accurately capture performance trends.

\subsection{Reinforcement Fine-Tuning with \ours{}}
\label{sec:expmathrl}

To investigate whether \ours{} models can be more effectively fine-tuned with RLVR, we randomly sample questions with verifiable answers from Skywork-OR1~\cite{skyworkor1} for further training. We use 256 examples for training and 200 for testing.
Following the data filtering pipeline from Skywork-OR1~\cite{skyworkor1}, we use R1-Distill-Qwen-32B to identify challenging instances for training.
We set both the training batch size and the PPO mini-batch size to 64, and train the model for 15 epochs.
During evaluation, the maximum number of tokens for validation is set to 32,000, with a temperature of 0.6.

As shown in \Cref{tab:mathrl}, the reinforcement pre-trained model achieves a higher upper bound when further trained with RLVR.
The reasoning ability of the model significantly declines when continually trained on the same data using a next-token prediction objective. Subsequent RLVR yields only slow performance improvements.
These results indicate that with limited data, \rpt{} can quickly transfer the strengthened reasoning patterns learned from next-token reasoning to end tasks.

\begin{table}[h!]
\centering
\label{tab:mathrl_accuracy}
\begin{tabular}{@{}lcc@{}}
\toprule
 & \bf Before RL & \bf After RL \\
\midrule
R1-Distill-Qwen-14B & 51.2 & 52.7\\
~~~~+ Continual NTP training & 10.7 & 13.0  \\
\ours{}-14B & \bf 56.3 & \bf 58.3 \\
\bottomrule
\end{tabular}
\vspace{5pt}
\caption{Reinforcement fine-tuning performance of different models.  ``Continual NTP training'' means continual pre-training using standard next-token prediction objective on the same corpus as \ours{}-14B. \ours{} provides a stronger foundation for subsequent RL training.}
\label{tab:mathrl}
\end{table}

\subsection{Zero-Shot Performance on End Tasks}
\label{sec:exptask}

We evaluate the zero-shot performance of \ours{}-14B on end tasks. 
For comparison, we assess the next-token prediction performance of R1-Distill-Qwen-14B and R1-Distill-Qwen-32B, as well as the the reasoning performance of \ours{}-14B with R1-Distill-Qwen-14B.

Our evaluation involves two widely acknowledged benchmarks: MMLU-Pro~\cite{mmlu}, a comprehensive multi-task understanding benchmark evaluating LLM capabilities across various domains; SuperGPQA~\cite{supergpqa}, a large-scale benchmark of graduate-level reasoning questions spanning 285 disciplines.
Under the reasoning setting, we set the maximum number of tokens to 12,288 and the temperature to 0.8. 
Following previous works~\cite{ma2025general,zhou2025reinforcing}, we use a multiple-choice question format for evaluation and report the accuracy.

\input{figure/all_general_tasks}

As shown in \Cref{tab:general-task}, \ours{}-14B consistently outperforms R1-Distill-Qwen-14B (whether using standard next-token prediction or evaluated as a reasoning model) across all benchmarks.
Notably, it also surpasses the significantly larger R1-Distill-Qwen-32B (under next-token prediction), with gains of 7 points on SuperGPQA and approximately 22 points on MMLU-Pro. Detailed per-subject results for each benchmark are provided in \Cref{app:details-endtask}.

\subsection{Next-Token Reasoning Pattern Analysis}
\label{sec:reasoning_pattern_analysis}

We analyze the differences in reasoning patterns between next-token reasoning and explicit problem solving. Following previous studies~\cite{Wang2025ReinforcementLF,rrm}, we statistically measure the proportion of model responses containing reasoning-indicative keywords (e.g., ``\textit{break down}'', ``\textit{alternatively}'').\footnote{The keywords are listed in \Cref{app:patterns}.}

Our analysis compares the thought processes of two models on the OmniMATH datasets, i.e., R1-Distill-Qwen-14B for problem solving, and \ours{}-14B for next-token reasoning, based on 200 sampled responses from each model.
We categorize reasoning patterns into six types: transition (switching strategies), reflection (self-checking), breakdown (decomposing the problem), hypothesis (proposing and verifying assumptions), divergent thinking (exploring possibilities), and deduction (logical inference).

\begin{figure}
\centering
\includegraphics[width=0.4\linewidth]{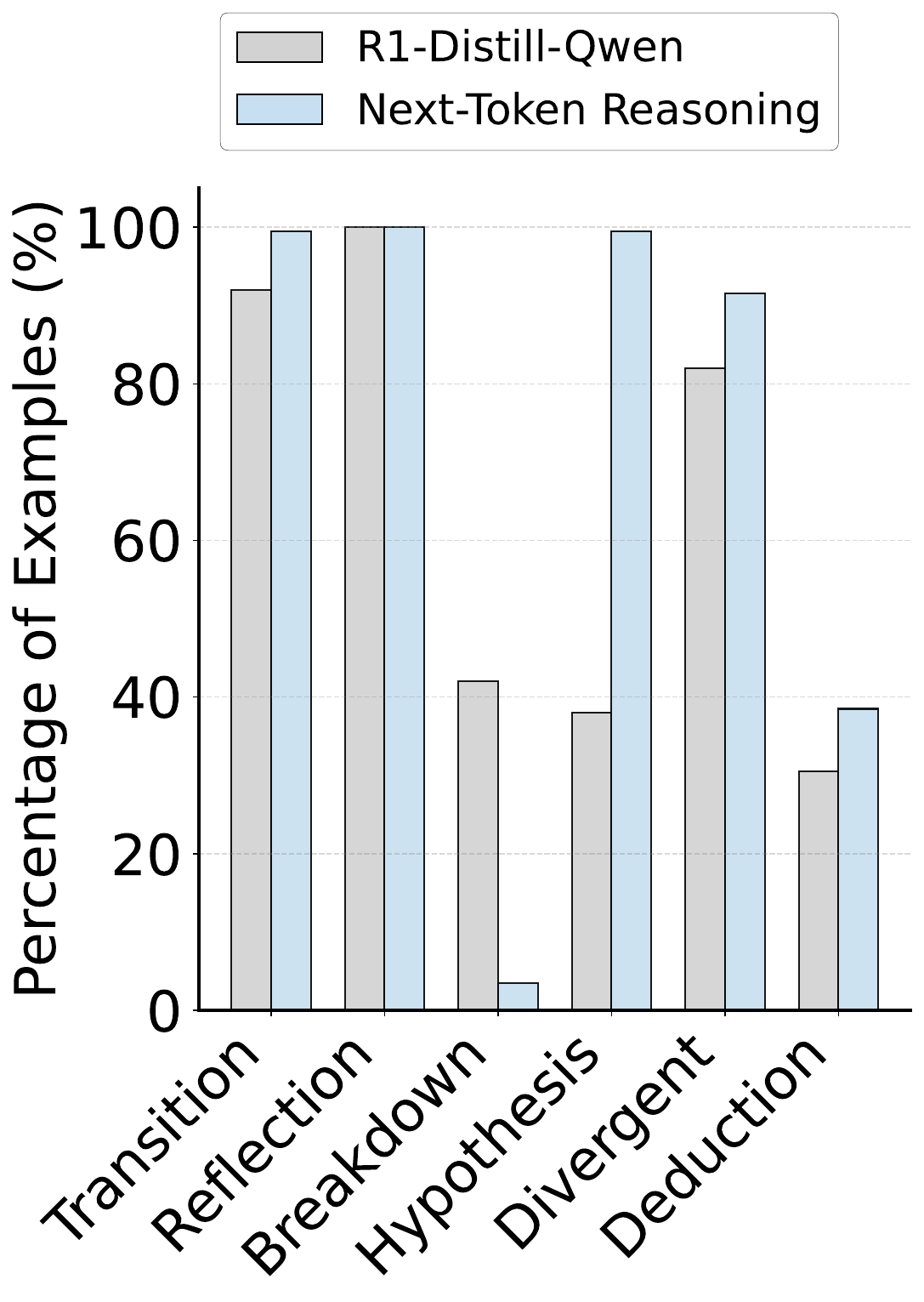}
\caption{Reasoning pattern statistics of R1-Distill-Qwen-14B employed for problem solving and \ours{}-14B employed for  next-token reasoning.}
\label{fig:pattern_groups}
\end{figure}

\begin{table}[t]
\centering
\scalebox{0.95}{
\begin{tabular}{p{14cm}}
\toprule
\textbf{Thinking Process of \ours{}-14B} \\
\toprule
Okay, I need to figure out what the next token ... Let's think about what would logically come next in a text about calculating vector magnitude. \newline
\textcolor{myblue}{The flow so far is} introducing the topic, explaining the formulas for 2D and 3D vectors, and now moving on to how to actually perform the calculation. Since the user mentioned ``go over some...'' \textcolor{myred}{the next part is likely going to be ...} \newline Alternatively, it could be ... \newline \textcolor{myred}{Common phrases after ``we can go over some'' could be ``examples,'' ``steps,'' ``methods,'' etc.} \textcolor{myblue}{Looking at similar contexts, in educational materials,} after explaining formulas, they often provide ``steps'' or ``examples.'' \newline But perhaps, given the user started a new section, \textcolor{myred}{it might be a new paragraph...} \newline \textcolor{myblue}{Wait, perhaps in the original, the next part was an example or an explanation}. Let's think about standard educational content. \newline So, the entire text might continue as: ... \newline Alternatively, perhaps the next line is a new section. \newline \textcolor{myblue}{Wait, another approach: ...} \newline However, the original text \textcolor{myblue}{could have a space before the next word...} \newline Wait, perhaps looking at how the initial document is structured. It's using markdown with headers:... \\
\bottomrule
\end{tabular}}
\vspace{5pt}
\caption{Example of reasoning patterns of \ours{}.}
\label{tab:case} 
\end{table}

As illustrated in \Cref{fig:pattern_groups}, \ours{}-14B's next-token reasoning process is markedly different from the problem-solving of R1-Distill-Qwen-14B, exhibiting a 161.8\% greater use of the hypothesis pattern and a 26.2\% greater use of the deduction pattern.
In contrast, the problem-solving process relies more heavily on the breakdown pattern, highlighting that next-token reasoning elicits an inferential process qualitatively different from structured problem-solving.

We also provide an example of reasoning patterns in \Cref{tab:case}.
The example reveals that the model engages in a deliberative process, not a simple pattern match. It analyzes the broader semantic context (``\textit{calculating vector magnitude}''), identifies pivotal phrases (``\textit{go over some...}''), and then brainstorms and weighs multiple plausible continuations.
This involves hypothesis generation (``\textit{the next part is likely going to be...}''), consideration of alternatives (``\textit{Alternatively, it could be...}''), and reflection on structural cues (``\textit{markdown with headers}'') and even fine-grained token-level details (``\textit{could have a space}''). This multi-faceted reasoning, encompassing both high-level semantic understanding and low-level textual features, demonstrates the model's effort to deduce the next token through a reasoned exploration, aligning with the goals of \ours{} to cultivate deeper understanding beyond superficial correlations.
More examples are provided in \Cref{app:case}.

\section{Related Work}

\paragraph{Scaling Paradigms of Large Language Models}
The advancements of large language models have been driven by two primary scaling dimensions: training-time compute~\cite{scaling:law,chinchilla} and test-time compute~\cite{survey-tts}.
Training-time scaling substantially increases model parameters and training data, using next-token prediction as the pre-training task.
Meanwhile, test-time scaling~\cite{o1} trades extended inference compute to improve the reasoning capabilities of large language models.
Going beyond existing scaling paradigms, \ours{} uniquely integrates the above principles, framing each next-token prediction as a reasoning task.

\paragraph{Reinforcement Learning for Large Language Models}
Reinforcement learning (RL) has played a crucial role in the post-training stage of large language models.
Reinforcement learning from human feedback \cite{instructgpt} fine-tunes pre-trained language models on human preference data to improve alignment.
Beyond alignment, large-scale RL has also been adopted to enhance the reasoning capabilities of language models~\cite{o1,deepseekr1}.
\cite{zelikman2024quiet} is the most relevant work, which encourages language models to generate helpful rationales for next-token prediction.
The helpfulness-based reward tends to be hacked by repeating the target token in the generated rationale, where the shortcut potentially harms the model.
In contrast, we use next-token prediction correctness as a rule-based reward signal to minimize reward hacking.

\section{Conclusion and Future Work}

We introduce \rpt{} (\ours{}), a novel paradigm for pre-training large language models. By framing next-token prediction as a verifiable reasoning task and applying reinforcement learning with correctness-based rewards, \ours{} allows LLMs to leverage extended computation during pre-training to build stronger foundational reasoning capabilities. Our experiments demonstrate that \ours{} improves next-token prediction, enhances performance on mathematical and general reasoning benchmarks in zero-shot settings, and provides a better starting point for further RL fine-tuning.
\ours{} offers a promising new direction for developing more capable and generally intelligent LLMs by fundamentally rethinking the pre-training objective itself.

While promising, this initial exploration of \ours{} has certain limitations. Our experiments are primarily conducted using a 14B parameter model. Although the \ours{} methodology is designed to be general, the current pre-training corpus predominantly consists of mathematical documents; future work will explore its efficacy on broader, general-domain text.
Furthermore, \ours{} training is initialized from a reasoning model; investigating \ours{} training from a standard base language model would provide further insights into its foundational impact.

The work can be advanced from the following perspectives.
We would like to scale up the training corpus, including data size, and domain coverage. Large-scale general Internet data can be utilized during \rpt{}.
We will also scale up training compute to push the frontier.
Moreover, we can establish scaling laws for \rpt{} to guide the scaling of large language models.
Additionally, we are interested in integrating hybrid thinking~\cite{hybridthink} with \ours{} to enable fine-grained adaptive thinking by adaptively triggering next-token reasoning.

\section*{Acknowledgement}

We extend our gratitude to Yuting Jiang for maintaining the GPU cluster.
We also thank Zewen Chi and Yang Wang for technical support during the development of the RL infrastructure on the MI300 GPUs.
We implement training based on verl~\cite{verl}.

\bibliography{rpt}
\bibliographystyle{alpha}

\newpage
\appendix

\section{Design Choices of Reward}
\label{app:reward}

We have also investigated several alternative reward functions to assess their impact on \rpt{}, in addition to the reward mechanism described in \Cref{sec:rpt}, i.e., prefix matching reward.

One variation is first-token matching. In this setup, the reward reflects only whether the first token of the model prediction $y_t^{i}$ matches the ground-truth next token $x_t$, ignoring all tokens after the first in the prediction.
Another alternative explores a `dense reward' scheme. Here, correctly predicted next tokens (i.e., $y_t^{i}[0] = x_t$) receive a full reward (e.g., 1). For incorrect predictions ($y_t^{i}[0] \ne x_t$), the reward is a positive, smaller value, specifically the language model probability of generating that particular incorrect token, $P(y_t^{i}[0] \mid x_{<t}; \theta)$. This provides a denser feedback signal than binary rewards.
A third design is a conditional application of this dense reward structure. The dense reward (1 for correct, $P(y_t^{i} \mid x_{<t}; \theta)$ for incorrect) is used as described above, but only for training instances (groups of rollouts for a given prefix $x_{<t}$) where at least one of the $G$ sampled rollouts correctly predicted the next token. If all $G$ rollouts in a group are incorrect, a different reward scheme (e.g., zero reward for all, or a uniform small penalty) will be applied.

Our experiments indicate that the alternative reward designs generally achieved performance comparable to the prefix matching reward. This suggests that the \rpt{} framework is relatively robust to these particular modifications in the reward signal, and its core benefits may not be overly sensitive to these specific choices, at least within the scope of variations tested.

\section{Hyperparameters Used for \RPT{}}
\label{app:hp}

\Cref{app:tbl:hp} presents the detailed hyperparameters for \rpt{} in \Cref{sec:exp}.
We follow the setting of exact on-policy reinforcement learning~\cite{opo} and set the entropy loss coefficient to $0$.

\begin{table}[h]
\centering
\begin{tabular}{lc}
\toprule
\textbf{Params} & \textbf{Values} \\
\midrule
Actor gradient clip & 0.2 \\
Batch size & {256} \\
PPO mini batch size & 256 \\
Rollout number & 8 \\
Learning rate & $10^{-6}$ \\
Adam $\beta$ & {(0.9, 0.999)} \\
Weight decay & 0.01 \\
Sampling temperature & 0.8 \\
Max prompt length & 4096 \\
Max response length & 8192 \\
Entropy loss coefficient & 0 \\
\bottomrule
\\
\end{tabular}
\caption{Hyperparamters used for \rpt{} in \Cref{sec:exp}.}
\label{app:tbl:hp}
\end{table}

\section{Detailed Results on End Tasks}
\label{app:details-endtask}

\Cref{tab:supergpqa} and \Cref{tab:mmlupro} present a detailed per-category performance across the general end task benchmarks. Notably, the performance of R1-Distill-Qwen-14B is evaluated in two different manner: standard next-token prediction and reasoning-based answer prediction (indicated as `+ think'). The \ours{}-14B model demonstrates superior performance compared to R1-Distill-Qwen-14B and R1-Distill-Qwen-32B.

\input{figure/supergpqa}
\input{figure/mmlupro}

\begin{table}[h!]
\centering
\begin{tabular}{lcc}
\toprule
Prompt Template & Random@1 (\%) & Pass@8 (\%) \\
\midrule
v0 & 3.0  & 8.5 \\
v1 & 5.7 & 11.0 \\
v2 & 5.7 & 16.0 \\
v3 & 5.3 & 11.0 \\
v4 & 4.0 & 9.0 \\
v5 & 4.4 & 12.5 \\
v6 & 6.0 & 19.0 \\
\bottomrule
\end{tabular}
\vspace{2.5pt}
\caption{Impact of prompt templates.}
\label{tab:prompt_template}
\end{table}

\begin{table}[h!]
    \centering
    \label{tab:pattern-keywords}
    \begin{tabular}{ll}
        \toprule
        \textbf{Pattern Group} & \textbf{Keywords} \\
        \midrule
        Transition & alternatively, think differently \\
        \midrule
        Reflection & wait, initial answer, original answer, looking back, thought process \\
        \midrule
        Breakdown & break down, break this down \\
        \midrule
        Hypothesis & probably, something like \\
        \midrule
        Divergent Thinking & etc., or something, either, sometimes it refers, otherwise, exploring, options \\
        \midrule
        Deduction & summarize, conclusion, conclude, finally, logically, consequently \\
        \bottomrule
    \end{tabular}
    \vspace{2.5pt}
    \caption{Pattern groups and keywords applied in ~\Cref{sec:reasoning_pattern_analysis}.}
    \label{tab:keywords}
\end{table}

\section{Impact of Prompt Templates}
\label{app:templates}

We explore the impact of various prompt templates on the initial next-token reasoning performance.
\Cref{tab:prompt_content} shows seven template variants.
The templates use different phrasing of instructions and wrap the context in various formats.

As presented in \Cref{tab:prompt_template}, clear prompts significantly improve the correctness of the initial performance.
Notice that the \rpt{} experiments in \Cref{sec:exp} used the `v0' prompt template.
We leave prompt engineering based on other template variants for future work, which tends to improve the final performance.

\input{figure/prompt_versions}

\section{Keywords for Reasoning Pattern Analysis}
\label{app:patterns}
\Cref{tab:keywords} presents the pattern groups and keywords applied in reasoning pattern analysis.

\section{Case Studies}
\label{app:case}

To provide qualitative insights into the behavior of models trained with \ours{}, we present some cases of next-token reasoning in \Cref{tab:three_cases}.

\begin{table}[htbp]
\centering
\scalebox{0.88}{
\begin{tabular}{p{14cm}}
\toprule

\textbf{Case 1} \\
\cmidrule{1-1}
Okay, I need to figure out what the next token is in the given text under the Context section. The user provided a partially completed text about integers, comparing and ordering them ... \textcolor{myblue}{First, I'll analyze the context provided} ... \newline
\textcolor{myred}{Wait, the user wrote 6 x 4 = 24, 20 > 18 x, so the next token is probably the second fraction} ... \newline
One way to find a common denominator is to multiply the two original denominators ... \newline
\textcolor{myblue}{Alternatively, it could be} ... \newline
Wait, but 18 x 24? That doesn't make sense. Alternatively, perhaps it's 18 x 4 = 72 or something... \newline
\textcolor{myred}{Another approach: perhaps the next token is} ... \newline
\textcolor{myblue}{Let me see:} if the two fractions are 5/3 and 4/6. 5/3 is greater than 4/6 because 5/3 is 1.666..., and 4/6 is 0.666..., to compare them, find a common denominator... \newline
Alternatively, \textcolor{myred}{perhaps the next token is a backslash or something in LaTeX}. Wait, but in the Markdown, the user presented it as: Writing Equivalent Fractions ... \newline
\textcolor{myblue}{But wait,} in the user's provided context, it's ... \newline
Maybe the next token is another multiplication symbol, but that seems unlikely. Alternatively, perhaps it's just the end of the thought. But given the context, \textcolor{myred}{I think that the next token is either a space followed by a '2' or a period or something else} ... \newline
\textcolor{myblue}{Wait, maybe another example.} If the fractions were 5/6 and 4/3 ... \newline
So, after 'x' there's a space, then '4'.\textbackslash n</think>\textbackslash n\textbackslash n\textcolor{myred}{So, the next token is ' 4'.\textbackslash n\textbackslash n\textbackslash boxed\{ 4\}} \\
\midrule
\textbf{Case 2} \\
\cmidrule{1-1}
Okay, I need to figure out what the next token is in the given text under the Context section. Using the integral test, for a set ... \newline
\textcolor{myblue}{Maybe the original answer was just a partial sentence. Let's think again.} The question is to prove convergence of the series, the answer uses the integral test, which is the standard approach ... \newline
Alternatively, maybe the original answer was longer, but got cut off, \textcolor{myred}{so I need to predict what comes next. Maybe the next token is} ... \newline
\textcolor{myred}{So the next token is a space and then something.} Alternatively, perhaps it's a continuation of the same sentence ... \newline
\textcolor{myblue}{Therefore}, by the integral test, the series converges when \$p > 1\$. No, perhaps it's better to see what the original answer was supposed to be. The correct full proof would be: Using the integral test, for a set ... \newline
\textcolor{myred}{But the user instruction is to provide the next token after the given text} ... \newline
\textcolor{myblue}{Alternatively}, looking at the way the answer was written, perhaps ... \newline
I think the next token is '\$'.\textbackslash n</think>\textbackslash n\textbackslash n\textcolor{myred}{So, the next token is '\$'.\textbackslash n\textbackslash n\textbackslash boxed\{\$\}} \\
\midrule
\textbf{Case 3} \\
\cmidrule{1-1}
Okay, I need to figure out what the next token is in the given text under the Context section. The text so far is: Call the large circle's center ... \newline
Scaling by a factor... \textcolor{myblue}{Perhaps the next part is the ratio of the radii? Or maybe something else. Wait}, the scaling factor is probably such that the smaller circle is scaled up to match the larger circle or something ...  \newline
So, in coordinates, perhaps O1 is at (0,0), O2 is at (41,0). The small circle has radius 13, so the points where O1O2 intersects the small circle are at O2 plus or minus 13 units towards O1 ... \newline
\textcolor{myblue}{So, point A is on the small circle, further away from O1 than O2, right.} Because if you go from O2 towards O1, 13 units, that would be ... \newline
\textcolor{myred}{So, the next token after 'Scale the small circle by a factor of' would probably be the number 30/13} ... \newline
Wait, I'm a bit confused. Alternatively, maybe it's presented as a fraction ... \newline
Scale the small circle by a factor of. \textcolor{myred}{So, the actual next token is the next word or symbol} ... \newline
Wait, but in the context given, the user just wrote ... \newline
\textcolor{myblue}{Alternatively, perhaps in the text it's written as} ... \newline
So, in the given text, the next token after 'Scale the small circle by a factor of' is space.\textbackslash n</think>\textbackslash n\textbackslash n\textcolor{myred}{The next token is ' '.\textbackslash n\textbackslash n\textbackslash boxed\{\ \}} \\
\bottomrule
\end{tabular}
}
\vspace{5pt}
\caption{Three examples of reasoning patterns of \ours{}-14B.}
\label{tab:three_cases}
\label{app:reasoning_case_study}
\end{table}

\end{document}

%% file: settings.tex
\usepackage{multirow}
\usepackage{amsmath}
\usepackage{capt-of}
\usepackage{tabularx}
\usepackage{epsfig}
\usepackage{amssymb}
\usepackage{amsfonts}
\usepackage{booktabs}
\usepackage{scalerel}
\usepackage[inline]{enumitem}
\usepackage{listings}
\usepackage{varwidth}
\usepackage[export]{adjustbox}
\usepackage{tikz}
\usetikzlibrary{tikzmark}

\usepackage{stmaryrd}
\usepackage{bbm}
\usepackage{wrapfig}
\usepackage{pifont}
\usepackage[noabbrev]{cleveref}

\definecolor{deepblue}{rgb}{0,0,0.5}
\definecolor{officeblue}{RGB}{0,102,204}
\definecolor{deepred}{rgb}{0.6,0,0}
\definecolor{deepgreen}{rgb}{0,0.5,0}
\definecolor{mybrickred}{RGB}{182,50,28}

\definecolor{fillcolor}{RGB}{216,217,252}

\usepackage{etoolbox}
\usepackage{framed}

\newif\ifxetexorluatex
\ifxetex
  \xetexorluatextrue
\else
  \ifluatex
    \xetexorluatextrue
  \else
    \xetexorluatexfalse
  \fi
\fi
\newcommand*\quotesize{60} 
\newcommand*{\openquote}
   {\tikz[remember picture,overlay,xshift=-4ex,yshift=-2.5ex]
   \node (OQ) {\fontsize{\quotesize}{\quotesize}\selectfont``};\kern0pt}

\newcommand*{\closequote}[1]
  {\tikz[remember picture,overlay,xshift=4ex,yshift={#1}]
   \node (CQ) {\fontsize{\quotesize}{\quotesize}\selectfont''};}

\colorlet{shadecolor}{white}

\newcommand*\shadedauthorformat{\emph} 

\newcommand*\authoralign[1]{%
  \if#1l
    \def\authorfill{}\def\quotefill{\hfill}
  \else
    \if#1r
      \def\authorfill{\hfill}\def\quotefill{}
    \else
      \if#1c
        \gdef\authorfill{\hfill}\def\quotefill{\hfill}
      \else\typeout{Invalid option}
      \fi
    \fi
  \fi}
{\authoralign{#1}
\ifblank{#2}
   {\def\shadequoteauthor{}\def\yshift{-2ex}\def\quotefill{\hfill}}
   {\def\shadequoteauthor{\par\authorfill\shadedauthorformat{#2}}\def\yshift{2ex}}
\begin{snugshade}\begin{quote}\openquote}
{\shadequoteauthor\quotefill\closequote{\yshift}\end{quote}\end{snugshade}}

\newfloat{Code}{htbp}{Code}

%% file: math_commands.tex
\usepackage{amsmath,amsfonts,bm}

\def\eqref#1{equation~\ref{#1}}

\def\1{\bm{1}}

\DeclareMathAlphabet{\mathsfit}{\encodingdefault}{\sfdefault}{m}{sl}
\SetMathAlphabet{\mathsfit}{bold}{\encodingdefault}{\sfdefault}{bx}{n}



%% file: figure/lm_ntp.tex
\begin{figure}[ht]
\centering
\begin{minipage}[c]{0.54\linewidth}
\centering
\setlength{\tabcolsep}{8pt}
\begin{tabular}{@{}lccc@{}}
\toprule
 & \bf Easy & \bf Medium & \bf Hard  \\
\midrule
\multicolumn{4}{l}{~~\textit{Standard next-token prediction}} \\
Qwen2.5-14B & 41.90 & 30.03 & 20.65 \\
R1-Distill-Qwen-14B & 41.60 & 29.46 & 20.43  \\
\midrule
\multicolumn{4}{l}{~~\textit{Next-token reasoning}} \\
R1-Distill-Qwen-14B  & 3.31 & 1.66 & 1.41\\
\ours{}-14B     & \bf 45.11 & \bf 33.56 & \bf 23.75  \\
\bottomrule
\end{tabular}
\captionof{table}{Next-token prediction accuracy across three test splits of varying difficulty. \ours{} outperforms both the standard next-token prediction baselines and the reasoning-based prediction baseline. }
\label{tab:ntp:acc}
\end{minipage}%
\hfill
\begin{minipage}[c]{0.42\linewidth}
\centering
\includegraphics[width=0.9\linewidth]{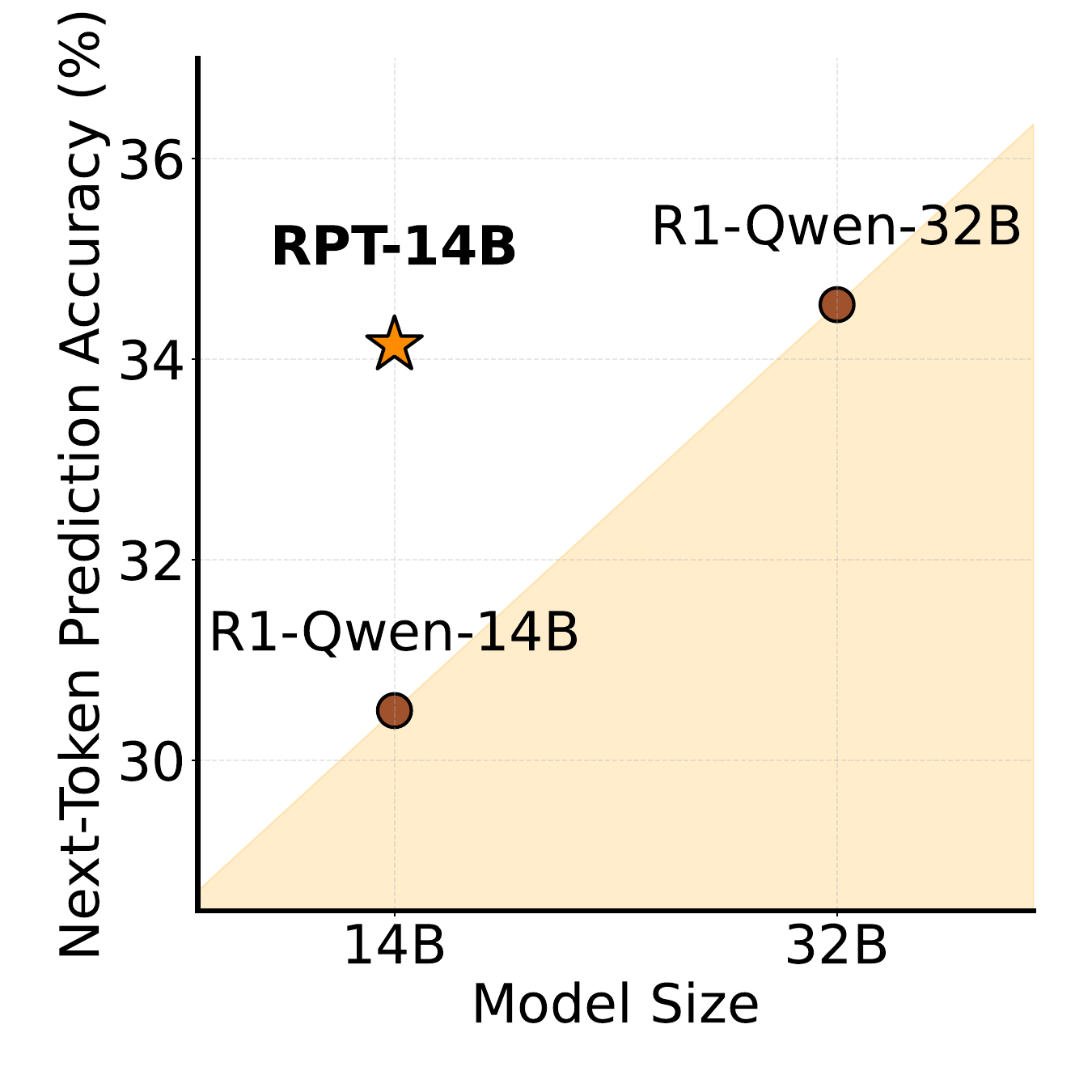}
\caption{Average next-token prediction accuracy across data of different difficulty levels. R1-Qwen-14B/32B denote R1-Distill-Qwen-14B/32B, respectively.
}
\label{fig:lm-avg}
\end{minipage}
\end{figure}

%% file: figure/all_general_tasks.tex
\begin{table}[htbp]
\centering
\setlength{\tabcolsep}{15pt}
\begin{tabular}{@{}lcc@{}}
\toprule
 & \textbf{SuperGPQA} & \textbf{MMLU-Pro} \\
\midrule
\multicolumn{3}{l}{\textit{Standard next-token prediction mode}} \\
R1-Distill-Qwen-14B & 32.0 & 48.4 \\
R1-Distill-Qwen-32B & 37.2 & 56.5 \\
\midrule
\multicolumn{3}{l}{\textit{Reasoning mode}} \\
R1-Distill-Qwen14B & 36.1 & 68.9 \\
\ours{}-14B        & \textbf{39.0} & \textbf{71.1} \\
\bottomrule
\end{tabular}
\vspace{5pt}
\caption{Zero-shot performance on general-domain end tasks. \ours{}-14B in reasoning mode consistently outperforms 14B and 32B baselines.}
\label{tab:general-task}
\end{table}

%% file: figure/supergpqa.tex
\begin{table}[htbp]
\centering
\setlength{\tabcolsep}{1pt}
\scalebox{0.85}{
\begin{tabular}{@{}lcccccccccccccc@{}}
\toprule
 & Agron. & Econ. & Educ. & Engin. & Hist. & Law & L.\&A. & Manag. & Med. & Mil. Sci. & Phil. & Sci. & Sociol. & Overall \\
\midrule
\multicolumn{3}{l}{~~\textit{Standard next-token prediction mode}} \\
R1-Distill-Qwen-14B & 30.0 & 38.0 & 32.0 & 31.0 & 24.5 & 26.0 & 28.5 & 39.0 & 35.5 & 36.0 & 37.0 & 24.0 & 30.1 & 32.0 \\
R1-Distill-Qwen-32B & 32.5 & 39.5 & 43.0 & 34.0 & 29.5 & 31.0 & 28.5 & 41.5 & 43.5 & 49.0 & 44.5 & 29.5 & 38.5 & 37.2 \\
\midrule
\multicolumn{8}{l}{~~\textit{Reasoning mode}} \\
R1-Distill-Qwen-14B & 31.0 & 41.0 & 32.0 & 34.5 & 29.0 & 31.0 & 29.5 & 39.5 & 38.5 & 39.5 & 44.0 & 41.5 & 39.2 & 36.1 \\
\ours{}-14B & 35.0 & 40.0 & 41.5 & 40.5 & 30.5 & 32.0 & 29.0 & 36.0 & 44.5 & 41.0 & 49.0 & 47.0 & 42.0 & 39.0 \\
\bottomrule
\end{tabular}
}
\vspace{2.5pt}
\caption{Detailed zero-shot performance on SuperGPQA.}
\label{tab:supergpqa}
\end{table}

%% file: figure/mmlupro.tex
\begin{table}[htbp]
\centering
\setlength{\tabcolsep}{1.55pt}
\scalebox{0.82}{
\begin{tabular}{@{}lcccccccccccccccc@{}}
\toprule
 & Bio. & Bus. & Chem. & CS & Econ. & Engin. & Heal. & Hist. & Law & Math & Other & Phil. & Phys. & Psych. & Overall \\
\midrule
\multicolumn{3}{l}{~~\textit{Standard next-token prediction mode}} \\
R1-Distill-Qwen-14B & 72.5 & 42.5 & 34.0 & 46.5 & 58.0 & 44.0 & 57.5 & 54.0 & 37.0 & 36.5 & 50.0 & 48.5 & 34.5 & 62.0 & 48.4 \\
R1-Distill-Qwen-32B & 82.5&46.0&39.0&55.5&74.0&52.0&68.0&62.5&47.0&46.0&54.0&53.5&42.5&68.5 & 56.5 \\
\midrule
\multicolumn{8}{l}{~~\textit{Reasoning mode}} \\
R1-Distill-Qwen14B & 85.0 & 65.5 & 74.5 & 75.0 & 81.5 & 52.0 & 70.0 & 61.5 & 42.0 & 86.0 & 65.0 & 62.5 & 80.0 & 64.5 & 68.9 \\
\ours{}-14B & 84.5 & 72.0 & 77.5 & 76.0 & 78.5 & 53.5 & 74.0 & 63.0 & 44.5 & 91.5 & 66.0 & 63.5 & 82.5 & 68.0 & 71.1 \\
\bottomrule
\end{tabular}
}
\vspace{2.5pt}
\caption{Detailed zero-shot performance on MMLU-Pro.}
\label{tab:mmlupro}
\end{table}

%% file: figure/prompt_versions.tex
\begin{table}[h!]
\centering
\small
\begin{tabular}{p{2cm}|p{12cm}}
\midrule
\textbf{Version} & \textbf{Prompt Content} \\
\midrule
v0 & Complete the given text under `\texttt{\#\#\# Context}' by predicting the next token, and wrap it in `\textbackslash boxed\{\}'.
Please reason step by step to find the most probable next token as the final answer, and enclose it in \textbackslash boxed\{\} (note: the token may begin with a space, e.g., \textbackslash boxed\{ para\} or \textbackslash boxed\{ =\}; do \textbf{not} use \textbackslash text\{\}).
\newline
\texttt{\#\#\# Context}
\newline
\{prompt\_content\} \\
\midrule
v1 & Complete the given text under \texttt{\#\#\# Context} by predicting the next token, and wrap it in \texttt{\textbackslash\textbackslash boxed\{\}}. Please reason step by step to find the most probable next token as the final prediction, and enclose it in \textbackslash boxed\{\} (note: the token may begin with a space, e.g., \textbackslash boxed\{ para\} or \textbackslash boxed\{ =\}; do \textbf{not} use \textbackslash text\{\}). \newline \texttt{\#\#\# Context} \newline \texttt{\textasciigrave \textasciigrave \textasciigrave\{prompt\_content\}\textasciigrave \textasciigrave \textasciigrave}. \\
\midrule

v2 & You are a helpful assistant, good at predicting the next token for a given context. \newline Now, please complete the given text under \texttt{\#\#\# Context} by predicting the next token, and wrap it in \texttt{\textbackslash\textbackslash boxed\{\}}. Please reason step by step to find the most probable next token, and enclose it in \textbackslash boxed\{\} (note: the token may begin with a space, e.g., \textbackslash boxed\{ para\} or \textbackslash boxed\{ +=\}; do \textbf{not} use \textbackslash text\{\}). \newline \texttt{\#\#\# Context} \newline \texttt{\textasciigrave \textasciigrave \textasciigrave\{prompt\_content\}\textasciigrave \textasciigrave \textasciigrave}. \\
\midrule

v3 & Complete the given text under \texttt{\#\#\# Context} by predicting the next token, list multiple potential tokens and select the most probable one as the final answer. Wrap your final answer in \texttt{\textbackslash boxed\{\}} (note: the token may begin with a space, e.g., \texttt{\textbackslash boxed\{ para\}} or \texttt{\textbackslash boxed\{ =\}}; do \textbf{not} use \texttt{\textbackslash text\{\}}). \newline \texttt{\#\#\# Context} \newline \texttt{\textasciigrave \textasciigrave \textasciigrave\{prompt\_content\}\textasciigrave \textasciigrave \textasciigrave} \\
\midrule

v4 & Complete the given text under \texttt{\#\#\# Context} by predicting the next token, and wrap it in \texttt{\textbackslash boxed\{\}}. Please reason step by step to find the most probable next token as the final answer, and enclose it in \textbackslash boxed\{\}. \newline Some examples: \newline \texttt{\#\#\# Context \textbackslash n \textbackslash n \textasciigrave \textasciigrave \textasciigrave...(some omitted)...Matching calculations with 1990 valid combinations indicates the minimum value of \textbackslash( b \textbackslash) that fits all pre-requisites and restrictions for triangle formation and symmetry generates the efficient outcome: \textbackslash n \textbackslash n \textbackslash[ \textbackslash n \textbackslash boxed\{1991\textasciicircum 2\} \textbackslash n \textbackslash] \textbackslash n \textbackslash nIn\textasciigrave \textasciigrave \textasciigrave} \newline The next token is \textbackslash boxed\{ this\} \newline \texttt{\#\#\# Context \textbackslash n \textbackslash n \textasciigrave \textasciigrave \textasciigrave...Thus \$2\textasciicircum \{A\}=\textbackslash left(2\textasciicircum \{a\}\textbackslash right)\textasciicircum \{2\}\textbackslash left(2\textasciicircum \{3\}
\textbackslash right)=\textasciigrave \textasciigrave \textasciigrave} \newline The next token is \textbackslash boxed\{9\} \newline \texttt{\#\#\# Context \textbackslash n \textbackslash n \textasciigrave \textasciigrave \textasciigrave..., numerical exploration shows\textasciigrave \textasciigrave \textasciigrave} \newline The next token is \textbackslash boxed\{:\textbackslash n\} \newline Now, the context is: \newline \texttt{\#\#\# Context \textbackslash n \textbackslash n \textasciigrave \textasciigrave \textasciigrave\{prompt\_content\}\textasciigrave \textasciigrave \textasciigrave}. \\
\midrule

v5 & Complete the given text under \texttt{\#\#\# Context} by predicting the next token, and wrap it in \texttt{\textbackslash boxed\{\}}. Please reason step by step to find the most probable next token as the final answer, and enclose it in \textbackslash boxed\{\} (note: the token may begin with a space, e.g., \textbackslash boxed\{ para\} or \textbackslash boxed\{ =\}; do \textbf{not} use \textbackslash text\{\}). \newline \texttt{\#\#\# Context} \newline \texttt{\textasciigrave \textasciigrave \textasciigrave\{prompt\_content\}\textasciigrave \textasciigrave \textasciigrave}. \\
\midrule

v6 & Complete the given text wrapped in \textasciigrave \textasciigrave \textasciigrave and \textasciigrave \textasciigrave \textasciigrave by predicting the next token, list multiple potential tokens and select the most probable one as the final prediction. Wrap your final prediction in \texttt{\textbackslash boxed\{\}} (note: the token may begin with a space, e.g., \texttt{\textbackslash boxed\{ para\}} or \texttt{\textbackslash boxed\{ =\}}; do \textbf{not} use \texttt{\textbackslash text\{\}}). \newline The context is: \texttt{\textasciigrave \textasciigrave \textasciigrave\{prompt\_content\}\textasciigrave \textasciigrave \textasciigrave}, now please predict the next token. \\
\midrule

\end{tabular}
\vspace{5pt}
\caption{Seven prompt templates for the next-token reasoning task.}
\label{tab:prompt_content}
\end{table}